# MHS-STMA: Multimodal Hate Speech Detection via Scalable Transformer-Based Multilevel Attention Framework

Anusha Chhabra*
*Department of Information Technology, Biometric Research Laboratory, Delhi Technological University,* Delhi-110042, India
anusha.chhabra@gmail.com

Dinesh Kumar Vishwakarma**
*Department of Information Technology, Biometric Research Laboratory, Delhi Technological University,* Delhi-110042, India
dinesh@dtu.ac.in

**Social media has a significant impact on people's lives. Hate speech on social media has emerged as one of society's most serious issues in recent years. Text and picture are two forms of multimodal data that are distributed within articles. Unimodal analysis has been the primary emphasis of earlier approaches. Additionally, when doing multimodal analysis, researchers neglect to preserve the distinctive qualities associated with each modality. To address these shortcomings, the present article suggests a scalable architecture for multimodal hate content detection called transformer-based multilevel attention (STMA). This architecture consists of three main parts: combined attention-based deep learning mechanism, a vision attention-mechanism encoder, and a caption attention-mechanism encoder. To identify hate content, each component uses various attention processes and handles multimodal data in a unique way. Several studies employing multiple assessment criteria on three hate speech datasets—Hateful memes, MultiOff, and MMHS150K— validate the suggested architecture's efficacy. The outcomes demonstrate that on all three datasets, the suggested strategy performs better than the baseline approaches.**

   **Keywords— Transformer, attention networks, hate speech, deep learning, multimodal analysis.**

## 1  INTRODUCTION

The emergence of social networking has facilitated people to exchange information quickly and easily, which has allowed for extensive communication in our daily lives [1]. However, social media has a dual use because it may also be used as a venue for spreading harmful content that may be misleading, offensive, or even extreme because of the anonymity it sometimes provides. Even with social media networks enforcing rules and conventions, it can be difficult to control offensive posts that contain malicious information. To lessen the impact of hate speech on online forums and in real-world situations, the identification of hate content within large amounts of social media information has become a hot topic. The task of detecting negative information from social media messages is difficult and complex. These online comments have the capacity to be quite hateful and could be impacted by the user's or a particular community's beliefs or opinions. Because social networks generate tremendous amounts of data every day, the process of content moderation is slowed down significantly when manual inspection is used as the primary method. This prolongs the time that offensive content is available on the internet. Memes have become an expected phenomenon on the internet in recent years. Memes are a combination of text and visual elements that can be found in a variety of formats, such as photographs and videos. Whatever the nature of the meme, it is typical for it to be altered and shared on different social media platforms in conversations about touchy themes like casteism and politics. Even though it can be difficult to interpret the underlying sentiment when there is text included in the photos, multimodal meme analysis can yield insightful information. This study's framework shows SOTA performance in removing offensive memes from social media data. The recognition of hate within extensive social media content has emerged as a prominent subject to mitigate the impact of such discourse within online platforms and real-life scenarios. The task of discerning nasty information from social media messages is difficult and complex. These online posts have the capacity to be extremely hateful and may be shaped by the user's or a particular community's prejudices or personal beliefs. Because social media platforms generate large amounts of data every day, the process of content moderation is slowed down significantly when human inspection is used as the primary method. This prolongs the time that offensive content is available on online platforms. Memes have become a popular phenomenon on the internet in recent times. Memes appear as a combination of text and visual elements since they are available in a variety of formats, such as photographs and videos. Whatever their origin, it's typical for memes to be altered and circulated on several social media platforms in conversations about touchy themes like casteism and politics. Even though it can be difficult to interpret the underlying sentiment when there is text included in an image, multimodal meme analysis can yield insightful information. The widespread use of social media platforms has led to the ongoing generation of uncontrolled data, which spreads unwanted content like hate speech and controversial opinions that incite violence. In recent years, hate speech has had a significant impact on the dynamics and importance of social media communications, generating alarm and international attention. Internet hate speech is gaining attention from the scientific community and policymakers since it is so prevalent on various social media and internet platforms. The pressing necessity to confront hate speech in its various forms and to guarantee equitable access to digital places is what is attracting this attention.

**\*\*(Corresponding Author)**

In this paper, we describe our scalable Transformer-based multilevel attention (STMA) framework, which consists of three primary components: combined attention-based deep learning mechanism, vision-attention-mechanism encoder for image branch, and caption-attention-mechanism encoder for textual branch. The first step is to add spatial information that corresponds to the various input image patches. Multihead self-attention (MSA), multilayer perceptron (MLP), and layer normalization are the methods used by the vision-attention-mechanism encoder to extract the abstract characteristics from the embedded patches. Contextual information is extracted from the input text sequence by the caption attention-mechanism encoder, which also creates a comprehensive embedding by merging token, segment, and position embeddings. Lastly, by choosing specific image features depending on the attended text features, the vision semantic attention block models the associations from the textual and visual data. The following is a summary of this article's major contributions:

1. Our proposal is an STMA framework that effectively models the interactions between textual and non-textual characteristics in multi-modal data by combining the strength of attention processes at multiple levels. The suggested technique successfully captures the semantic connections between the textual and visual characteristics by a cross-attention mechanism.
2. Additionally, we provide the multihead attention (MHA) mechanism, which integrates data from various attention levels. To be more precise, the framework would employ several heads of attention to handle various components of the multimodal data. This would enable a broad variety of interactions between the textual and non-textual characteristics to be captured by the framework.
3. Using three publicly available datasets, we assess the performance of the proposed method using a variety of metrics and visualization tools. The outcomes are compared to other state-of-the-art methods. Additionally, we prepared and verified every step of the procedure by carrying out the ablation investigation.

The article is arranged as follows for the remainder of it. In Section II, the relevant research on hate speech identification is covered. The suggested architecture is explained in Section III. In Section IV, we use various assessment criteria to confirm our model's functionality. Section V wraps off by talking about the current work's future direction and conclusion.

## 2 RELATED WORK

Several cutting-edge methods for detecting hateful content based on a single modality have been covered in this area. Numerous studies have been conducted to categorize user-generated textual content on social media sites in relation to offensive language and hate speech. For quicker automatic hate speech identification, [2] suggested adding parallelization to a regular ensemble learning model made up of many ML classifiers. [3] utilised an ensemble model including many machine learning classifiers to detect hate speech in tweets pertaining to COVID-19. [4] used a heterogeneous stacking-bagging technique to create an ensemble model with a variety of base learners, including CNN, LSTM, BiLSTM, and BiGRU, to attain better results on multilingual hate speech detection datasets. [5] examined the results of utilizing several embedding methods in combination with ML classifiers to detect offensive and hateful content in Tamil literature. [6] created a BERT-based trained model for the identification of hate. They also investigated the possibilities of using prior activity analysis and the detection of hate content to determine whether a social media user profile is being managed by a hater. [7] suggested textual hate speech detection in multilingual datasets by means of transfer learning via pre-trained cross-lingual language models. [8] presented the unique EHSor framework in the context of multi-label learning, which enhances hate speech detection by relying on emotion states. In order to enhance the identification of hate speech in languages with limited resources, [9] suggested utilizing a transfer learning approach that relies on pre-trained cross-lingual language models. In order to reveal model bias toward keywords and offer solutions, the researchers in their study [10] contrasted the most significant terms identified by transformer models optimized for hate speech recognition with a list of hateful keywords taken from the datasets. [11] presented the use of low-parameter character-level hypernetworks for enhanced hate speech detection. The authors also applied dataset augmentation via text generation to further boost the performance of their proposed model. [12] utilized a BERT layer in conjunction with a hierarchical attention module and BiLSTM network to segregate hateful tweets. Through their experimentation in [13] demonstrated how machine translation and a pretrained English language model could help achieve good performance for the identification of hateful expressions in low-resource languages. [14] curated an Urdu language hate speech detection dataset and made use of machine learning classifiers as well as transformer models to conduct baseline experiments on the same. In order to improve outcomes in the field of hate speech identification, [15] suggested using pertinent data from related classification texts, such as abusive language, aggression, and harassment detection. The authors of the paper [16] examined the generalizability of textual hate speech classification models across several datasets and determined which essential model and dataset attributes were necessary for cross-dataset generalization. The authors in [17] created a probabilistic clustering approach for hate speech classification since binary hate speech classifiers ignore the emotions that overlap between the positive and negative classes. [18] combined complementing characteristics derived from various ML feature extraction approaches to create a multi-classifier system for efficient hate speech identification. [19] demonstrated that, in contrast to black-box deep learning networks like transformers, machine learning techniques based on text embeddings and fuzzy rough sets offered a more comprehensible framework for hate speech detection. [20] looked into how well hate speech recognition models worked in extracting common traits from datasets that were independent of topics and using that information to recognize hate speech. [21] presented a methodology based on computer vision to identify unsuitable and non-compliant product logos and pictures. [22] have used skin detection algorithms to recognize offensive content—specifically, nudity—in photos and videos. [23] increased the accuracy of identifying pornographic photos by taking advantage of the representational capacities of several models. The authors proposed using a fusion technique for prediction, which makes use of the knowledge from various transfer learning models. To detect pornography, [24] integrated the low- and mid-level features of many state-of-the-art pre-trained models. Furthermore, the GGOI dataset for obscene picture recognition was made available by the authors. Extensive research on a particular modality, such text or pictures, has been conducted on the full examination of hate speech and objectionable content. Nonetheless, the combination of two modalities to identify offensive information is still an emerging field of research. By subtly expressing irony and sarcasm, multimodal inputs give another level of complexity [25], which may lessen the offensiveness compared to textual or visual analyses alone. Therefore, in order to assess the degree of offensiveness displayed by a specific meme, it is imperative that both modalities be taken into consideration. In order to combine textual and visual elements for automatic multimodal hate speech detection, [26] experimented with the fusion technique. In a multimodal approach, [27] employed text, pictures, and OCR to identify antisemitism in two datasets gathered through Twitter and Gab. To gain insights into the context and relationship between the two modalities in hateful memes, [30] investigated by employing sentiment analysis and pre-trained image captioning models. [31] curated a video dataset for the purpose of hate speech detection and thereby introduced a multimodal framework that effectively combines acoustic features related to emotion with semantic features to identify hateful content accurately.

## 2 PROPOSED METHODOLOGY
This section describes the proposed architecture in detail.

### 3.1 Problem Definition
A set of multimodal samples $M = \{m_1, m_2, \ldots, m_n\}$ is given, where each $m_i \in M$ has an image $I_i$ with the corresponding target $T_i$ and captions with $w_i$ words. Attached to each $T_i$ is a label $y_i$, which may be hate or no-hate. To achieve a uniform distribution of both modalities, we first eliminated those cases from the datasets that contain either caption or image data. Images and text undergo different preprocessing steps. The natural language toolkit (NLTK) package is used to preprocess text input. It assists in eliminating stop words and stemming and lemmatizing words to return them to their root form. Images are scaled and their mean is subtracted to achieve normalization. In addition, we have employed several data augmentation methods such as flipping, rotating, zooming, and so on to prevent the model from being overfit to the training set. Using the proposed STMA framework, our aim is to predict the proper label for the collection of unseen samples.

### 2.2 Patch Embeddings
Every image $I_i$ is separated into smaller patches, and each one makes use of a $16 \times 16$ convolution with a stride of 16. The fixed-size patches from the batch of input photos with the shapes $(b, h, w, and\ c)$ are flattened to create the flat patches. We apply a trainable embedding vector of dimension d to these patches. This provides us with a linear embedding of the flattened patches in low dimensions. To obtain a consolidated representation of all the patches, a learnable token is prepended to the patch embeddings. After that, we include the positional embeddings so that the transformer model is fully aware of the image sequence. We are adding the spatial data associated with every patch in the series in this way.

### 2.3 Vision Attention-Mechanism Encoder
The transformer attention-based encoder receives the patched embeddings produced in Section-III (B) and uses them to learn the abstract features. We have employed the Vision transformer as the foundational framework for the visual data. The following elements are essentially included in the encoder module: layer normalization (Norm), MLP, and MSA. Self-attention has the advantage of being able to extract information from the full visual globally. Consequently, the MSA block splits the inputs into numerous heads, each of which is capable of learning and comprehending the various facets of the input's abstract representation. All the heads' output is combined and sent to the MLP layer, which makes use of the GeLu nonlinearity. To cut down on the amount of time the network needs to train, layer normalization is applied before each layer. Additionally, residual connections are used to get around the issue of the vanishing gradient.

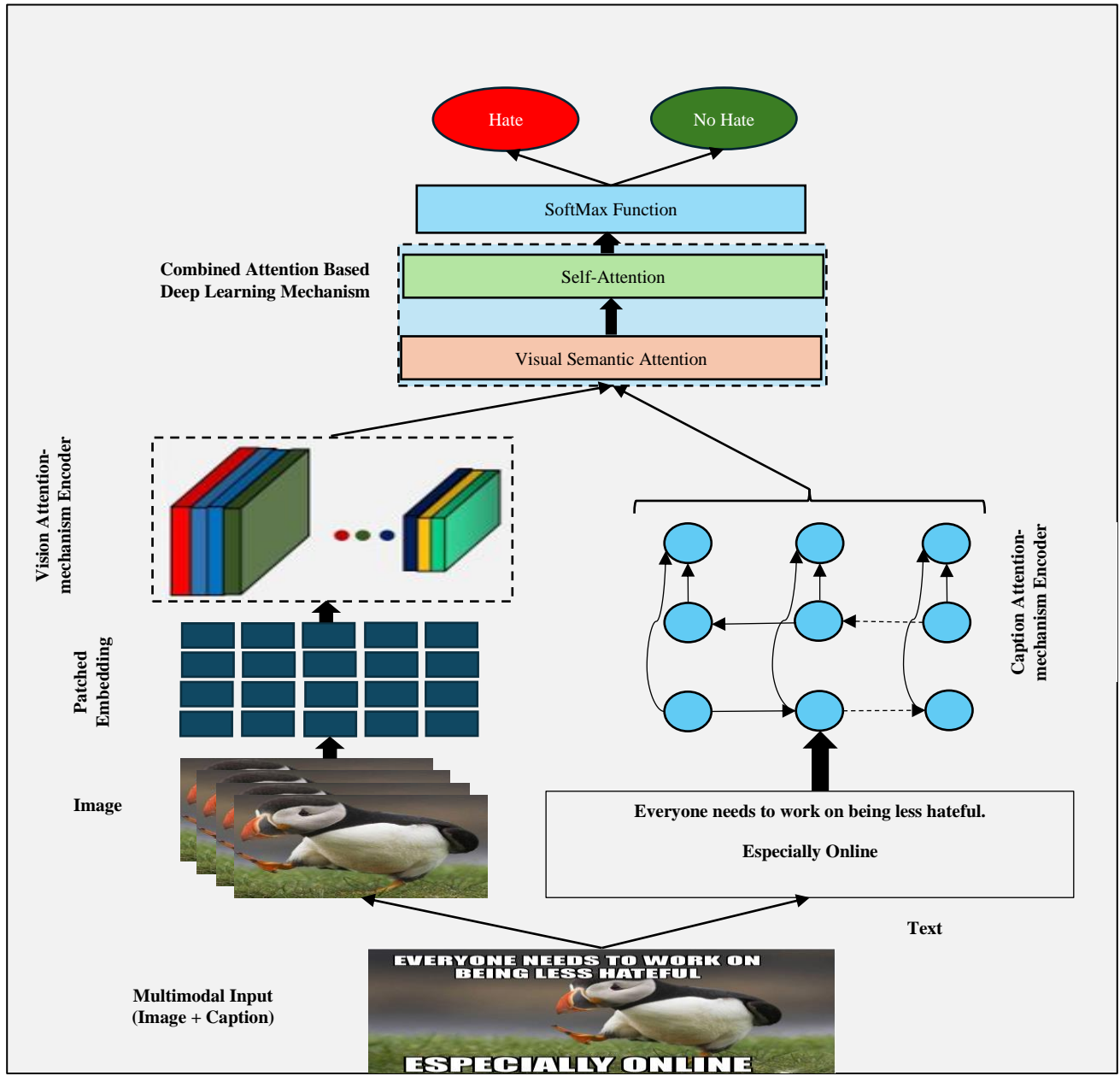

Figure 1. Proposed STMA Framework

## 2.4 Text Attention-Mechanism Encoder

The bidirectional encoder (BERT) representation from Transformers [28] is used to encode the raw text sequences, once more making use of the attention mechanism. Token embeddings, segment embeddings, and position embeddings are combined to turn the text sequences into tokens. Token embeddings (Ti) provide the vocabulary IDs for each token, sentence embeddings (Si) aid in sentence differentiation, and position embeddings (Pi) show the word positions inside sentences. Every embedding layer is linked to the sublayers before it and comprises distinct MSA sublayers. The discriminative characteristics separating the text and image modalities are not learned by the multimodal analysis works now in use. It becomes essential to investigate the complementing information between the various modalities in multimodal feature learning. This will improve our model's overall performance.

## 3.5 Combined Attention-Based Deep Learning

Two modules are used in combined attention-based deep learning mechanism to accomplish this. Initial module is the visual semantic attention block, which creates multimodal features by extracting important picture aspects from attended text information. A self-attention block in the second module eliminates features from the multimodal data that aren't needed.

- The goal of the **visual semantic attention block** is to understand which image features to prioritize, using the words in the caption sequence. The visual semantic attention block receives an image-caption pair $\{I_i, C_i\}$ for the ith sample. Element-wise multiplication is utilized to combine two modalities to achieve this.
- Several modalities collaborate in the **self-attention block** to determine which feature should be prioritized and to calculate the attention of all the inputs in relation to one another. This is crucial since merging the modalities could produce a lot of unrelated features. The interaction between the multimodal elements—which include both text and image features—allows for the identification of the features that require additional attention. Because of this, the self-attention block will combine the attention of all the inputs with respect to one another, highlighting the various multimodal features based on their weights.

Finally, the SoftMax classifier receives the final features acquired and uses them for classification. A probabilistic activation function is called SoftMax. For every output label, it provides the likelihood that the label belongs to the class. The output chosen for the final class is the one with the highest probability.

*Table 1*
*Algorithm for the proposed STMA Architecture*

| Algorithm 1: Multimodal Hate Speech Detection via Scalable Transformer-Based Multilevel Attention Framework |
|---|
| **Input:** Set of multi-modal samples $M = \{m_1, m_2, \ldots, m_n\}$. Each $m_i \in M$ contains captions with $w_i$ words and an image $I_i$ with an associated target $T_i$. Each $T_i$ is attached with label $l_i$. |
| **Output:** Hate Content Classification task as **hate, or no-hate** |
| 1. **Patch Embedding:**<br>• Split image $I_i$ into patches of $16 * 16$ convolution having stride 16.<br>• To generate the embedding, multiply with the embedding vector.<br>• Add positional embedding to create the patched embedding.<br>2. **Vision Attention-mechanism Encoder**<br>• To understand the input's abstract representation, divide the input patches into several heads.<br>• Combine all head outputs and pass them to the MLP layer which contains one hidden and an output layer.<br>3. **Caption Attention-mechanism Encoder**<br>For a sequence of 'n' words:<br>Encode the captions sequence by token, sentence and position embedding as:<br>$$E(f_i) = \{T_i + S_i + P_i\} \forall\ i = 1,2,\ldots,n$$<br>4. **Combined Attention-based deep Learning mechanism**<br>• Pass the multimodal sample into the block of visual semantic attention.<br>• Use self-attention to eliminate any characteristics that are unnecessary.<br>• Utilize the SoftMax classifier to categorize the input sample as either hateful or not.<br>5. **End** |

## 4 Experimental analysis

This section outlines the experimental conditions, and the procedures used to determine the suggested framework's level of proficiency.

### 4.1 Dataset Description

The following multimodal datasets were utilized to test and train the suggested framework in order to see how well it could identify nasty memes.

#### 4.1.1 Multi Modal Hate Speech Dataset (MMHS150K):

A multimodal hate speech dataset of 150,000 tweets was generated in [32] and named MMHS150K. The collection contains textual information and supporting images for every tweet. We used the Twitter API to gather tweets in real time. The authors removed the tweets that contained textual images in order to ensure that all dataset instances included both textual and visual information.

#### 4.1.2 Multimodal Meme Dataset for Offensive Content (MultiOff):

Using the 2016 U.S. Presidential Election as a point of reference for identifying objectionable content on social media, the authors in [33] created a multimodal dataset with 743 memes that were divided into offensive and non-offensive classifications.

#### 4.1.3 Hateful Memes Challenge (HMC):

[34] presented a difficult dataset for identifying hate speech in memes. Because of the way the dataset is structured, only multimodal frameworks can effectively classify the memes, with unimodal techniques unable to do so. This is accomplished

by adding confounding samples to the collection, which makes depending just on one modality challenging.

### 4.2 Hyperparameters Settings
**Table 2** provides the specifics of the experimental hyperparameter settings for each dataset, including the number of epochs, batch size, starting learning rate, and optimizer.

*Table 2 Hyperparameters Settings*

| Dataset | Number of Epochs | Batch Size | Learning Rate | Optimizer |
|---|---|---|---|---|
| MMHS150K | 10 | 32 | 0.0001 | Adam |
| MultiOff | 40 | 8 | 0.001 | Adam |
| HMC | 25 | 16 | 0.0001 | Adam |

### 4.3 Data Pre-Processing
The pre-processing techniques used in the current experiment are described in this section. Every image has its dimensions set to a standard 3 x 256 x 256. After normalizing, the pixel values are in the range [0,1].

### 4.4 Train, Validation, and Test Splits
This section contains the total number of samples in each of three datasets. The ratio of training, validation and testing sets is 8:1:1, respectively is shown in **Table 3**.

*Table 3 Dataset size (total, training, testing and validation)*

| Dataset | Size | Training Set | Validation Set | Testing Set |
|---|---|---|---|---|
| MMHS150K | 150000 | 120000 | 15000 | 15000 |
| MultiOff | 743 | 600 | 70 | 70 |
| HMC | 8496 | 6800 | 840 | 840 |

Two NVIDIA TITAN RTX GPUs with a combined memory capacity of 24 GB are used in the research. Both GPUs run simultaneously.

## 5 Results and Discussion
This section presents a performance and comparison analysis of the results obtained.

### 5.1 Performance and Comparison against SOTA on Benchmark Datasets
This section presents the outcomes of the proposed architecture on the MultiOff, HMC, and MMHS150K datasets. **Table 4** shows the figures for accuracy, precision, recall, F1 score, and area under the curve in addition to a comparison with SOTA techniques. The enhanced multimodal hate speech detection approach that has been suggested effectively extracts crucial data from both textual and visual modalities. The MultiOff, HMC, and MMHS150K datasets yielded accuracy scores of 0.6509, 0.8790, and 0.8088, respectively, indicating a notable enhancement in performance. The AUC scores of 0.6857, 0.8500, and 0.7840 also demonstrate a noteworthy improvement in performance when compared to previous studies.

*Table 4 Performance and Comparison*

| Dataset | Ref | Acc | P | R | F1 | AUC |
|---|---|---|---|---|---|---|
| MultiOff | [33] | - | 0.4000 | 0.6600 | 0.5000 | - |
| | [35] | - | 0.6450 | 0.6510 | 0.6480 | - |
| | **Ours** | **0.6509** | **0.6740** | **0.6940** | **0.6839** | **0.6857** |
| Hateful Memes | [34] | 0.6947 | - | - | - | 0.7544 |
| | [35] | 0.7580 | - | - | - | 0.8280 |
| | [36] | 0.7650 | - | - | - | 0.8374 |
| | [37] | 0.7108 | 0.7000 | - | 0.6900 | 0.7141 |
| | **Ours** | **0.8790** | **0.8348** | **0.6140** | **0.7678** | **0.8500** |
| MMHS150K | [32] | 0.6850 | - | - | 0.7040 | 0.7340 |
| | [38] | 0.7143 | - | - | 0.7085 | - |
| | [39] | - | - | - | - | 0.7149 |
| | [40] | - | 0.6133 | 0.5134 | 0.5589 | - |

| | Ref | Acc | P | R | F1 | AUC |
|---|---|---|---|---|---|---|
| | [41] | 0.7401 | - | - | - | 0.7634 |
| | **Ours** | **0.8088** | **0.7108** | **0.7388** | **0.7246** | **0.7840** |

## 5.2 Ablation Trials

To examine the impact of the individual components in our suggested architecture, we do ablation research in this part. We do the multi-modal analysis on all the datasets after first conducting the uni-modal analysis on the caption and vision data independently. **Table 5** provides a summary of the findings.

### 5.2.1 Uni-modal Analysis

The caption input goes through a caption attention-mechanism encoder, which is then followed by self-attention for the caption modality. The features that have been extracted are sent to the softmax layer for the last stage of classification. For the visual aspect, we create patched embeddings and send them to the visual attention-mechanism encoder module, then implementing the self-attention mechanism. The ultimate characteristics are passed straight to the softmax classifier. In both scenarios, the visual semantic attention block is removed because we are working with unimodal data exclusively.

### 5.2.2 Multi-modal Analysis

In multi-modal analysis, we assess the importance of each component by removing different elements from our proposed framework. The visual semantic block's multimodal features are sent to the softmax classifier without considering the self-attention block. Afterwards, we remove the visual-semantic attention block from the architecture, considering both the self-attention block and softmax layer.

The significance of integrating the semantic correlation between visual and caption features is evident in **Table 5**. Next, the vision attention-mechanism encoder block is taken out, the patched embeddings are sent through the pretrained VGG-16 model, and combined-attention based deep learning mechanism is carried out. The findings clearly confirm the significance of our vision attention-focused encoder block in capturing the unchanged characteristics of the images. Ultimately, we disable the encoder that focuses on captions and observe that attention to the captions plays a vital role, as it highlights key words and assists in setting the context.

*Table 5* Ablation Scores

| | Model | Accuracy | | |
|---|---|---|---|---|
| | | **MultiOff** | **Hateful Memes** | **MMHS150K** |
| **Unimodal** | Textual | 0.5667 | 0.6585 | 0.7437 |
| | Visual | 0.5333 | 0.3750 | 0.7500 |
| **Multimodal** | Without Visual Semantic Attention | 0.5989 | 0.6900 | 0.7689 |
| | Without Self Attention | 0.5764 | 0.6756 | 0.7490 |
| | Without Vision Attention-mechanism encoder | 0.6091 | 0.7501 | 0.7736 |
| | Without Caption Attention-mechanism encoder | 0.6117 | 0.7607 | 0.8025 |
| | **Proposed** | **0.6509** | **0.8790** | **0.8088** |

## 6 Qualitative Visualization

Memes' captions and visual portions both include a substantial quantity of information that is undeniable. The informative portion of the image is represented by the spatial region. It locates the pertinent visual components based on the visually attended elements. The activation mapping via attention approach, i.e., GradCAM,[42] is visualized in Table 6 to find the fine-grained localization of objects. GradCAM requires a gradient to be present on a given layer to capture the target layer's attention.

*Table 6* Spatio-Region of Importance via GradCAM

| | Memes | Focused Region |
|---|---|---|
| MultiOff | | |
| Hateful Memes | | |
| MMHS150K | | |

The observations from **Table 6** are:
- Improving object localization by concentrating on the designated spatial region.
- The GradCAM activation map ascertains the influence of each region on a model's output.

## 7    Conclusion and Future work

Social media platforms have enabled diverse modes of communication, hence enabling a comprehensive and swift interchange of ideas. Millions of people utilize these platforms, and among them are active participants in the posts that are shared. Even with the inclusion of social norms and procedures on these platforms, it is still difficult to stop the spread of hateful and unwanted postings. It is a difficult task to identify nasty information from multimodal posts. These posts could be overtly hateful, or they might be shaped by the individual beliefs of a specific user or group. Reliance on human evaluation slows down the procedure and increases the possibility that the offensive material will stay up online for a long time. As a result, it is imperative to set up efficient technologies that can identify offensive content on social networking sites without requiring human assistance. This research provides a revolutionary multimodal framework that is capable of efficiently removing hostile memes. The effectiveness of the recommended methodology is demonstrated by the ease with which the suggested architecture beats the current baselines. The dearth of scholarly works delving into the topic of multimodal hate content identification is indicative of the vast array of untapped research prospects. We are inspired to apply the suggested architecture's outstanding performance to additional well-known multimodal domains including sentiment analysis, sarcasm detection, and fake news identification.